\documentclass[journal]{IEEEtran}
\usepackage{cite}

\ifCLASSINFOpdf
\else
\fi
\usepackage{graphicx}
\usepackage{float}
\usepackage{multirow}
\usepackage{stfloats}
\usepackage{makecell}
\usepackage{amsmath,amssymb,amsfonts}
\usepackage{caption}
\usepackage{subfig}
\usepackage{color}
\usepackage{algorithm}
\usepackage{algpseudocode}
\usepackage{algorithmicx}
\usepackage{makecell}
\usepackage{color}
\usepackage{booktabs}
\pdfoutput=1

\begin{document}
%
\title{Learning Domain-invariant Graph for Adaptive Semi-supervised Domain Adaptation with Few Labeled Source Samples}
%
%
%

\author{Jinfeng~Li, Weifeng~Liu,~\IEEEmembership{Senior Member,~IEEE,}
        Yicong~Zhou,~\IEEEmembership{Senior Member,~IEEE,}
        Jun~Yu,~\IEEEmembership{Senior Member,~IEEE,} 
        Dapeng~Tao,~\IEEEmembership{Senior Member,~IEEE,}}

%
%

\markboth{IEEE TRANSACTIONS ON KNOWLEDGE AND DATA ENGINEERING,~Vol.~, No.~, JUNE~2020}%
{Shell \MakeLowercase{\textit{et al.}}: Bare Demo of IEEEtran.cls for IEEE Journals}
%



\maketitle

\begin{abstract}
 Domain adaptation aims to generalize a model from a source domain to tackle tasks in a related but different target domain. Traditional domain adaptation algorithms assume that enough labeled data, which are treated as the prior knowledge are available in the source domain. However, these algorithms will be infeasible when only a few labeled data exist in the source domain, and thus the performance decreases significantly. To address this challenge, we propose a Domain-invariant Graph Learning (DGL) approach for domain adaptation with only a few labeled source samples. Firstly, DGL introduces the Nystr$\rm\ddot{o}$m method to construct a plastic graph that shares similar geometric property as the target domain. And then, DGL flexibly employs the Nystr$\rm\ddot{o}$m approximation error to measure the divergence between plastic graph and source graph to formalize the distribution mismatch from the geometric perspective. Through minimizing the approximation error, DGL learns a domain-invariant geometric graph to bridge the source and target domains. Finally, we integrate the learned domain-invariant graph with the semi-supervised learning and further propose an adaptive semi-supervised model to handle the cross-domain problems. The results of extensive experiments on popular datasets verify the superiority of DGL, especially when only a few labeled source samples are available.
\end{abstract}


%
\IEEEpeerreviewmaketitle

\section{Introduction}
%
%
%
%

\IEEEPARstart{I}{n} the big data era, large amounts of data with distribution divergence emerge in various application scenarios. It is unrealistic for the huge time-consuming and labor-costing to label massive samples for a specific task. Moreover, traditional machine learning methods are workable only under the assumption that training data (source domain) and testing data (target domain) are independent and identically distributed (i.i.d), and rich labeled data are needed during the training period~\cite{PanA,DBLP:journals/corr/abs-1903-04687}. Therefore, it is becoming a great challenge to analyze newly-generated data characterized by scarce label information and distribution divergence. Recently, domain adaptation that can effectively narrow down the distribution gap between source and target data~\cite{Sinno2011Domain,MingshengTransfer,YanLearning,WeiA} attracts lots of attention, and it has been applied to a wide range of applications, e.g., image categorization~\cite{Hua2011Dyadic,LeiDomain,DingRobust} text classification~\cite{Zhuang2011Mining,article,Li2017Knowledge}, and so on.

The source domain and target domain involved in domain adaptation are usually generated from different but related probability distributions. Thus, domain adaptation proposes to reduce distribution divergence between the two domains, so that a model established in the source domain can be well generalized to the target domain. Recently, a line of domain adaptation frameworks have been proposed and can be classified into two categories summarily~\cite{PanA}, including instance reweighting adaptation~\cite{DBLP:conf/icml/DaiYXY07,DBLP:journals/pami/ChuTC17,DBLP:conf/vcip/ChenZL16} and feature representation adaptation~\cite{long2015domain,DBLP:conf/icann/DasL18,DBLP:journals/corr/CourtyFTR15}.

The methods based on instance reweighting strategy generally assign different weights to source samples to reduce distribution differences across domains so that the models built in the source domain can be adaptive to the target domain. Dai \emph{et al.}~\cite{DBLP:conf/icml/DaiYXY07} proposed the TrAdaBoost framework under the hypothesis that certain parts of source data are still useful while the remaining are outdated. It iteratively calculates the weights of source data with the help of a few labeled target data and reuses source data with larger weights and a few labeled target data to train a classifier. Chu \emph{et al.}~\cite{DBLP:journals/pami/ChuTC17} proposed the Selective Transfer Machine (STM) method to reduce domain-specific mismatches by simultaneously learning a generic classifier and reweighting the source data related to target subject. Chen \emph{et al.}~\cite{DBLP:conf/vcip/ChenZL16} presented a weighted subspace learning method, which aligns two subspaces respectively generated from the reweighted source data and the target data for across domain adaptation.

The feature representation adaptation methods aim to find a good feature representation of different domains. Long \emph{et al.}~\cite{long2015domain} introduced the Transfer Kernel Learning (TKL) method based on Nystr$\rm\ddot{o}$m approximation error minimization to learn a cross-domain kernel.
 Das \emph{et al.}~\cite{DBLP:conf/icann/DasL18}  used a graph matching metric to measure the domain discrepancy. It realizes a domain-invariant and category-discriminative representation by minimizing this metric. Courty \emph{et al.}~\cite{DBLP:journals/corr/CourtyFTR15} proposed an unsupervised optimal transportation work. It learns a local transportation map for each source sample, so that source samples are close to target samples.

  Although many domain adaptation algorithms have been reported to connect source and target domains via different strategies, almost all domain adaptation algorithms mentioned above focus on the situation where enough labeled source data can be available. However, it is expensive to collect massive labeled source data or annotate newly-emerged data in real applications. As a result, these algorithms will fail to bridge two domains without enough labeled data existing in the source domain, and the performance reduces significantly. To address this challenge, we propose the Domain-invariant Graph Learning (DGL) method to implement knowledge transfer across domains with only a few labeled source samples. Specifically, DGL introduces the  Nystr$\rm\ddot{o}$m method~\cite{DBLP:conf/icml/ZhangTK08} to construct a plastic graph ${\bar{L}}^s$ which shares the similar geometric property as target domain. Then, DGL utilizes Nystr$\rm\ddot{o}$m approximation error to measure the distance between the plastic graph ${\bar{L}}^s$ and the source graph $L^s$ to formalize the distribution mismatch between domains from the geometric perspective. Through minimizing the cross-domain approximation error, DGL can learn a domain-invariant geometric graph ${\bar{L}}^{s\ast}$ to link source and target domains. In particular, the invariant geometric graph can facilitate DGL to employ unlabeled source data to assist knowledge transfer and thus reducing the reliance on source label information. Finally, we integrate the learned domain-invariant graph ${\bar{L}}^{s\ast}$ with the semi-supervised learning and further propose an adaptive semi-supervised model. In addition, we extend RLS and SVM such two standard methods under the adaptive model to effectively handle the cross-domain classification problems.

 The main contributions of DGL can be summarized as follows:

\begin{itemize}
\item We propose a Domain-invariant Graph Learning (DGL) approach to tackle the challenge of domain adaptation with only a few labeled source samples.
\item DGL flexibly uses the Nystr$\rm\ddot{o}$m method to learn a domain-invariant graph to link source and target domains from the geometric perspective. To our best knowledge, it is the first attempt to transfer learning with such a strategy.
\item We integrate integrate the learned invariant graph with the semi-supervised framework and further propose an adaptive semi-supervised model to decrease the dependence on source domain label information during the transfer process.
\item Many widely-used standard methods, e.g., RLS and SVM can be easily integrated with the proposed adaptive model to handle the cross-domain classification problems.
\item We construct comprehensive experiments on different kinds of datasets to demonstrate the superiority of DGL, especially when scarce labeled source samples are available.
\end{itemize}

The subsequent sections are organized as follows. We introduce the related works of DGL in Section ~\ref{section2}. And then, we detail the DGL algorithm and further proposed an adaptive cross-domain model in Section~\ref{section3}. In section~\ref{section4}, we extend two widely-used classification algorithms, i.e., RLS and SVM under the adaptive model with different choices of loss function to deal with the cross-domain classification problems. In Section~\ref{section5}, we conduct experiments on different kinds of datasets to evaluate the performance of DGL. We conclude the paper in in Section~\ref{section6}.

\section{Related Work}\label{section2}
In this section, we discuss some related works to DGL including geometric adaptation and the Nystr$\rm\ddot{o}$m method.
\subsection{ Geometric Adaptation}
Geometric adaptation methods involve a hypothesis that input samples are sampled from a low-dimensional manifold structure, and similar samples should be encoded with similar representations~\cite{belkin2006manifold,MingshengTransfer}. It aims to better link source and target domains by exploiting the low-dimensional structure which reflects the intrinsic distribution of input samples.

Gong \emph{et al.}~\cite{DBLP:conf/cvpr/GongSSG12} proposed a geometric flow kernel (GFK) method to extract the infinite number of subspaces that change along the geodesic flow across domains. Yao \emph{et al.}~\cite{DBLP:conf/cvpr/YaoPNLM15} proposed the Semi-supervised Domain Adaptation with Subspace Learning (SDASL) method to find a subspace where the divergence between the source and target mappings of the same class is minimized. And it uses manifold regularization to exploit the intrinsic information of the target domain. Wang \emph{et al.}~\cite{wang2018visual} proposed the Manifold Embedded Distribution Alignment (MEDA). It introduces manifold feature learning to exploit the geometric property of the data in Grassmann manifold to construct a cross-domain classifier better.

The major restriction of these domain adaptation works is that they seriously rely on a large amount of source label information, which is labor-consuming to collect in real applications. Therefore, we propose DGL to deal with one more practical problem of domain adaptation where only a few labeled source samples are available.

 \subsection{Nystr$\ddot{\textbf{o}}$m Method}
The Nystr$\rm\ddot{o}$m method is initially introduced to seek the solution of integral Eq.\ref{1} ~\cite{DBLP:conf/icml/ZhangTK08,1262185}.
\begin{equation}\label{1}
\begin{split}
\int p(t) \phi_{i}(t) k(s, t) d t=\lambda_{i} \phi_{i}(s)
\end{split}
\end{equation}
where $k(\cdot)$  is a function which is positive semidefinite and square integrable.  $p(\cdot)$ is probability density function.  $\lambda_{i}$ and $\phi_{i}(\cdot)$ are eigenvalues and eigenfunctions of integral Eq.\ref{1}.

In the theory of kernel machine analysis, the Nystr$\rm\ddot{o}$m method is employed to approximate the Gram matrix to reduce the computational burden~\cite{williams2000using,drineas2005on,li2015large-scale}. Given a dataset $\mathrm{T}=\left\{t_{i}\right\}_{i=1}^{n_{1}}$ sampled from the distribution $p(t)$, a kernel matrix $K_{T}$ built based on $\mathrm{T}$. Given another new dataset $\mathrm{S}=\left\{s_{i}\right\}_{i=1}^{n_{2}}$  sampled from \textbf{the same distribution} of $\mathrm{T}$ . The eigenfunction  $\phi_{i}(s)$ on new instance $s_{i}$  can be estimated by Eq. \ref{2}:
\begin{equation}\label{2}
\begin{split}
\phi_{i}(s) \simeq \sum_{j=1}^{n_{1}} \frac{k\left(s, t_{j}\right) \phi_{i}\left(t_{j}\right)}{n_{1} \lambda_{i}}
\end{split}
\end{equation}

The eigenfunction $\phi_{i}(s)$  on new dataset $\mathrm{S}$ is approximated in a matrix form as follows:
\begin{equation}\label{3}
\begin{split}
\Phi_{S} \simeq K_{S, T} \Phi_{T} \Lambda_{T}^{-1}
\end{split}
\end{equation}
where $\Lambda_{T}=\operatorname{diag}\left(\lambda_{1}, \ldots, \lambda_{n_{1}}\right) \in R^{n_{1} \times n_{1}}$  are  $n_{1}$ eigenvalues and $\Phi_{T} \in R^{n_{1} \times n_{1}}$  are $n_{1}$  eigenvectors of $K_{T}$, i.e., $K_{T}=\Phi_{T} \Lambda_{T} \Phi_{T}^{T}$. $K_{S, T} \in R^{n_{2} \times n_{1}}$  is the similarity matrix between two datasets. $\Phi_{S} \in R^{n_{2} \times n_{1}}$  is the extrapolated eigenvector matrix of kernel matrix $K_{S}$ . The kernel matrix $K_{S}$  can be estimated by Eq.\ref{4}:
\begin{equation}\label{4}
\begin{split}
K_{S} \simeq \Phi_{S} \Lambda_{T} \Phi_{S}^{T}=K_{S, T}\left(\Phi_{T} \Lambda_{T}^{-1} \Phi_{T}^{T}\right) K_{T S}=K_{S, T} K_{T}^{-1} K_{T, S}
\end{split}
\end{equation}

Attracted by the properties and successful applications of the Nystr$\rm\ddot{o}$m method, we flexibly apply it to domain adaptation and use it to discuss geometric relationships between domains. By flexibly employing the Nystr$\rm\ddot{o}$m method, we build a domain-invariant graph to bridge two domains geometrically. To our best knowledge, it is the first attempt for transfer learning with such a strategy.
\section{Domain-invariant Graph Learning}\label{section3}
In this section, we detail the DGL algorithm including problem formulation, general framework, learning optimization, and complexity analysis.
\subsection{Problem Formulation}
Suppose there are $n_t$ samples $X_t=\left\{x_i^t\right\}_{i=1}^{n_t}\in R^{m\times n_t}$ in target domain $D_t$, and $n_s$ samples $X_s=\left\{x_i^s\right\}_{i=1}^{n_s}\in R^{m\times n_s}$ in source domain $D_s$. Specifically, source domain includes $u$ unlabeled samples $X_t^u=\left\{x_i^{su}\right\}_{i=1}^u\in R^{m\times u}$ and $l$ labeled samples $X_s^l=\left\{x_i^{sl}\right\}_{i=1}^l\in R^{m\times l}$ $(u\gg l,u+l=n_s)$. Denote $Y_s$ as the source label matrix. In DGL, only scarce source label information can be available, referring to $Y_s^l{=\left\{y_i^{sl}\right\}}_{i=1}^l$. Generally, the data of two domains are generated from the identical $m$ dimension feature space with the same $C$ classes. But they have different marginal probability distribution, i.e., $\mathcal{P}_s\neq\mathcal{P}_t$.

Let ${{L}^{\centerdot }} $  represents a source or target graph Laplacian ( $\centerdot $ can be either source or target) and ${{\Phi }^{\centerdot }} $ and ${{\Lambda }^{\centerdot }} $ represent the eigenvector and eigenvalue matrices of graph Laplacian ${{L}^{\centerdot }} $ respectively.
In the following sections, we compute the Laplacian matrix as $L^{\centerdot }=D^{\centerdot }-W^{\centerdot }$, where $W^{\centerdot }$ is the affinity matrix given by Gaussian kernel $k$ with the form $W_{ij}^{\centerdot }=\exp \left( {-\left\| x_{i}^{\centerdot }-x_{j}^{\centerdot } \right\|_{2}^{2}}/{2{{\sigma }^{2}}}\; \right) $. $\sigma$ is bandwidth and fixed 1 in this paper. $D^{\centerdot }$ is the diagonal matrix computed by $D_{ij}^{\centerdot }=\sum_{j=1}^{\ }W_{ij}^{\centerdot }$.

\subsection{ General Framework}
DGL is proposed to address a new problem of domain adaptation where only a few labeled source samples are available. Specifically, DGL geometrically links the source and target domains via a domain-invariant graph Laplacian ${{\bar{L}}^{s*}}$. More specifically, DGL uses the Nystr$\rm\ddot{o}$m method to build a plastic graph ${\bar{L}}^s$, which has the similar geometric property of target graph. Then DGL utilizes the Nystr$\rm\ddot{o}$m approximation error to measure the divergence between the plastic graph ${\bar{L}}^s$ and source graph $L^s$ to formalize the distribution mismatch. Finally, the domain-invariant graph can be successfully learnt via minimizing Nystr$\rm\ddot{o}$m method approximation error and naturally integrated into the semi-supervised framework based on source data with the promising performance for target data.
\subsubsection{Learning a domain-invariant graph}
Source and target data following similar marginal distribution naturally requires that both domains have similar geometric structure, i.e., $L^s\backsimeq L^t$. Considering that the graph Laplacian matrices of different domains generally have different dimensions, i.e., $L^s\in R^{n_s\times n_s}$, $L^t\in R^{n_t\times n_t}$. It is impossible to measure the discrepancy between different Laplacian matrices directly. Therefore, we use the Nystr$\rm\ddot{o}$m method~\cite{DBLP:conf/icml/ZhangTK08} flexibly to tackle this problem by generating a plastic graph Laplacian $
\bar{L}^s\in R^{n_s\times n_s}
$ based on target graph Laplacian eigenvector matrix $\Phi ^t$. According to the Nystr$\rm\ddot{o}$m method, $\bar{L}^s$ has the similar geometric property of of target graph $L^t$. Therefore, $\bar{L}^s$ can replace $L^t$  to compare with $L^s$.

Firstly, we  apply eigen-decomposition on target graph Laplacian $L^t$, thus we get the eigensystem of target graph structure $\left\{ \Phi ^t,\Lambda ^t \right\}$.
\begin{equation}\label{6}
\begin{split}
L^t=\Phi ^t\Lambda ^t\left( \Phi ^t \right) ^T
\end{split}
\end{equation}

According to the  Nystr$\rm\ddot{o}$m method, we can derive the estimated eigenvector matrix of source graph as follows:
\begin{equation}\label{7}
\begin{split}
{\bar{\Phi}}^s\simeq L^{st}\Phi ^t\left( \Lambda ^t \right) ^{-1}
\end{split}
\end{equation}
where $L^{st}$ is the cross-domain graph, which can be computed by $L^{st}=L\left( 1:n_s,n_s+1:n_s+n_t \right)$. $L$ is the graph Laplacian on all data. It can be computed by $L=D-W$, $W_{ij}=exp\left( \dfrac{-\lVert x_{i}-x_{j} \rVert _{2}^{2}}{2\sigma ^2} \right)$ and $D_{ii}=\sum_{j=1}{W_{ij}}$.

In the Nystr$\rm\ddot{o}$m method introduced in Section~\ref{section2}, the source graph can be directly approximated by ${\bar{\Phi}}^s$ and $\Lambda ^t$ similarly to Eq.~\ref{4}. However, Eq.~\ref{4} can be established if and only if two domain data have the same distribution, which obviously does not hold in domain adaptation. In other word, when $\mathcal{P}_t\ne \mathcal{P}_s$, the source graph achieved by the Nystr$\rm\ddot{o}$m method on target domain is not accurate and will cause large approximation error. Nevertheless, this approximation error makes us realize that Nystr$\rm\ddot{o}$m approximation error embodies distribution divergence and can be used to formalize the domain difference, i.e., when $\mathcal{P}_s\approx \,\,\mathcal{P}_t$ the error is approximately zero~\cite{long2015domain}. Therefore, by minimizing the Nystr$\rm\ddot{o}$m approximation error we can find a geometric graph that is naturally domain-invariant and can be used to bridge two domains.

Specifically, we relax target graph eigenvalue matrix $\Lambda ^t$ to be learnable variable $\Lambda$ and build a plastic graph $\bar{L}^s$ based on $\left\{ {\bar{\Phi}}^s,\Lambda ^{\,\,} \right\}$ using the Nystr$\rm\ddot{o}$m method, i.e., $\bar{L}^s={\bar{\Phi}}^s\Lambda \left( {\bar{\Phi}}^s \right) ^T$. Note that because ${\bar{L}}^s$ is extrapolated from target graph eigensystem, therfore, ${\bar{L}}^s$ naturally has the similar geometric property of target graph $L^t$. In contrast, the source graph $L^s$ is built based on source data and captures the geometric characteristics of source data. Therefore, the Nystr$\rm\ddot{o}$m approximation error can adequately represent the distribution divergence.

Then we have the minimization problem of the Nystr$\rm\ddot{o}$m approximation error between the plastic graph Laplacian ${\bar{L}}^s$ and the source graph Laplacian $L^s$ as follows:
\begin{equation}\label{9}
\begin{split}
min _{\Lambda} & \lVert \bar{L}^s-L^s \rVert _{F}^{2}=\lVert {\bar{\Phi}}^s\Lambda \left( {\bar{\Phi}}^s \right) ^T-L^s \rVert _{F}^{2} \\
 & s.t. \ \lambda _i\,\,\geq 0,\,\,i=1,...,n_t,\ \\
& \lambda _i\geq \,\,\xi \lambda _{i+1},\ \,\,i=1,...,\,\,n_t-1
\end{split}
\end{equation}
where $\Lambda =\mathrm{diag}\left( \lambda _1,...,\lambda _{n_t} \right) $ are learnable free eigenvalues. $\xi \geq 1$ is eigenspectrum damping factor, and it can enable the larger eigenvectors to make more contributions for knowledge transfer~\cite{long2015domain}.

Reformulate Eq.~\ref{9} into a matrix form:
\begin{equation}\label{10}
\begin{split}
min_{\lambda}\lambda ^T\mathcal{Q}\lambda -2\mathcal{R}^T\lambda \\
s.t.\,\mathcal{Z}\lambda \geq 0,\lambda \geq 0
\end{split}
\end{equation}
$\mathcal{Q}$, $\mathcal{R}$, and $\mathcal{Z}$ are defined as follows:
\begin{equation}\label{11}
\begin{array}{c}
\mathcal{Q}=\left(\left(\bar{\Phi}^{s}\right)^{T} \bar{\Phi}^{s}\right) \odot\left(\left(\bar{\Phi}^{s}\right)^{T} \bar{\Phi}^{s}\right) \\
\mathcal{R}=\operatorname{diag}\left(\left(\bar{\Phi}^{s}\right)^{T} L^{s} \bar{\Phi}^{s}\right) \\
\mathcal{Z}=I-\xi \bar{I}
\end{array}
\end{equation}
where $\lambda=\left(\lambda_1,\ ...\lambda_{n_t}\right)$, $I\in R^{n_t\times n_t}$  is identity matrix, $\bar{I}\in R^{n_t\times n_t}$ is a matrix with nonzero elements $\bar{I}_{i,i+1}=1,i=1,...,n_t-1$. The objective~\ref{10} is a QP probelm with linear constrains, it can be easily solved by many convex optimization packages~\cite{inbook}.

Solving the problem~\ref{10}, we get the eigenvalue matrix $\Lambda ^* $. Then the domain-invariant geometric graph is computed as follows:
\begin{equation}\label{12}
\begin{split}
\bar{L}^{s*}={\bar{\Phi}}^s\Lambda ^*\left( {\bar{\Phi}}^s \right) ^T
\end{split}
\end{equation}
where $\Lambda ^* $ is learned eigenvalue matrix by optimizing ~\ref{10}.

The domain-invariant geometric graph $\bar{L}^{s*}$ is constructed based on eigensystem $\left\{ {\bar{\Phi}}^s,\Lambda ^*\,\, \right\}$. Specifically, it preserves the main structure of target graph $L^t$ via the eigenvector matrix ${\bar{\Phi}}^s$.  Simultaneously it can minimize the distribution difference via eigenvector matrix $\Lambda ^*$, which is determined across domains. Thus, the learnt $\bar{L}^{s*}$ is an appropriate bridge linking source and target domain geometrically.
\subsubsection{Adaptive Semi-supervised Model}
 Particularly, we use DGL to learn an adaptive cross-domain model $f$. Considering the scarce source label information, we adopt the idea of semi-supervised learning, which employs manifold regularization~\cite{belkin2006manifold} to take advantage of a great deal of unlabeled source data. Integrating $\bar{L}^{s*}$ with manifold regularization, we further propose an adaptive semi-supervised model for target domain.
\begin{equation}\label{15}
\begin{split}
f=argmin_{f\in \mathcal{H}_k}\frac{1}{l}V\left( x_{i}^{sl},y_{i}^{sl},f \right) +\lambda _1\lVert f \rVert _{k}^{2}+\lambda _2\mathbf{f}^T\bar{L}^{s\ast}\mathbf{f}
\end{split}
\end{equation}
where $\mathcal{H}_k$ represents a Reproducing Kernel Hilbert Space (RKHS), $k$ is the kernel function. $V\left(\ast\right)$ is the loss function defined on labeled source samples. $x_i^{sl}$ represents labeled source sample with the corresponding label $y_i^{sl}$, $\lambda_1$ is the regularization parameter controlling the complexity of the adaptive model and $\lambda_2$ is another regularization parameter controlling the geometric adaptation. $\mathbf{f}=\left[f\left(x_{1}^{s}\right), \ldots, f\left(x_{l}^{s l}\right), f\left(x_{1}^{s u}\right) ,\ldots, f\left(x_{u}^{s u}\right)\right]^{T}$. The term $\mathbf{f}^T{\bar{L}}^{s\ast}\mathbf{f}$ is an appropriate penalty item that reflects the shared geometric properties across domains.

The model $f$ induced by Eq.~\ref{15} can be used to predict the target data. Notably, on the one hand, $\bar{L}^{s*}$ captures the common geometric properties across domains. On the other hand, $\bar{L}^{s*}$ involves all source data, which makes the unlabeled source data can assist knowledge transfer. Therefore, ${\bar{L}}^{s\ast}$ is the main point, which enables DGL can transfer knowledge across domains without relying on much source label information.

\section{Adaptive cross-domain classifier construction}\label{section4}
Through discussions in Section~\ref{section3}, we obtain an adaptive semi-supervised model. Notably, $V\left(\ast\right)$ is the most flexible part of the proposed model. Without loss of generality, we integrate two widely-used loss functions, including the hinge loss for SVM $V\left(\ast\right)=max\left(0,1-y_if\left(x_i\right)\right)$ and the squared loss for RLS $V\left(\ast\right)=\left(y_i-f\left(x_i\right)\right)^2$, into the proposed model~\ref{15} to construct two adaptive cross-domain classifiers.

Suppose the adaptive cross-domain classifier be $f=\omega^T\psi\left(x\right)$, where $\omega$ is the classifier parameters, and $\psi$ is the mapping function that projects the original feature into the RKHS $\mathcal{H}_k$.

To construct the classifier effectively, we resort to use the ``Representer Theorem''~\cite{belkin2006manifold} to solve the optimization problem~\ref{15} under different choices of $V\left(\ast\right)$. The classical ``Representer Theorem'' shows that the solution to the~\ref{15} can be written as:
\begin{equation}\label{16}
\begin{split}
f\left( x \right) =\sum_{i=1}^{l+u}{\alpha _i}k\left( x,x_i \right)
\end{split}
\end{equation}
Specifically, $\omega =\sum_{i=1}^{l+u}{\alpha _i\psi \left( x_i \right)}$. $k$ is the kernel function induced by $\psi$ such that $k\left( x,x_i \right) =\left< \psi \left( x \right) ,\psi \left( x_i \right) \right>$. The optimization problem~\ref{15} is reduced to obtain the coefficients $\alpha_i$.

In this paper, we construct the RLS and SVM under the adaptive semi-supervised model with the corresponding loss function $V\left(\ast\right)$ and use the ``Representer Theorem'' to solve the optimal problems effectively. The details are discussed in following two parts.
\subsection{Adaptive RLS Classifier}
Using the squared loss $V\left( f\left( x_{i}^{sl} \right) ,y_{i}^{sl} \right) =\left( y_{i}^{sl}-f\left( x_{i}^{sl} \right) \right) ^2$ and the ``Representer Theorem'',  the objective~\ref{15} based on RLS can be formulated as follows:
\begin{equation}\label{17}
\begin{aligned}
\boldsymbol{\alpha}=\operatorname{argmin}_{\boldsymbol{\alpha} \in R^{(l+u)}} \frac{1}{l}\left\|\left(Y_{s}-\boldsymbol{\alpha}^{T} K\right) J\right\|_{F}^{2}+\\
\operatorname{tr}\left(\lambda_{1} \boldsymbol{\alpha}^{T} K \boldsymbol{\alpha}+\lambda_{2} \boldsymbol{\alpha}^{T} K \bar{L}^{s *} K \boldsymbol{\alpha}\right)
\end{aligned}
\end{equation}
where $\boldsymbol{\alpha}=\left( \alpha _1,...,\alpha _{n_s} \right)^T$  is the coefficient matrix. $J$ is a diagonal matrix given by $J=diag\left(1,...,1,0,...,0\right)\in R^{n_s\times n_s}$ with the first $l$ diagonal elements as 1 and the rest 0. $Y_s$ is the label matrix on source data. It does not matter that some source label information is unknown, because they can be filtered out by the label indicator matrix $J$.
 $K\in R^{\left(l+u\right)\times\left(l+u\right)}$ is the Gram matrix over all source data. To simplify calculations, we choose Gaussian kernel function $k$ to compute the Gram matrix in this paper.

The derivative of the objective function is set to \textbf{0} and we can obtain:
\begin{equation}\label{18}
\begin{split}
\boldsymbol{\alpha}=\left(\left(J+\lambda_{2} l \bar{L}^{s *}\right) K+\lambda_{1} l I\right)^{-1} J Y_{s}^{T}
\end{split}
\end{equation}

Denote $y\in R^C$ as label vector, then $y_c=1$ if $y\left(x\right)=c$ and otherwise $y_c=0$. The label matrix $Y_s=\left[y_1^s,...,y_{n_s}^s\right]\in R^{C\times n_s}$, and the parameter matrix $\boldsymbol{\alpha}\in R^{n_s\times C}$. In such a way, the adaptive classifier based on DGL can be used to deal with the multi-class problem.

\subsection{Adaptive SVM Classifier}
Using the hinge loss: $V\left(f\left(x_{i}^{s l}\right), y_{i}^{s l}\right)=\max \left(0,1-y_{i}^{s l} f\left(x_{i}^{s l}\right)\right)$ and the ``Representer Theorem'', the objective~\ref{15} based on SVM can be formulated as follows:
\begin{equation}\label{19}
\begin{array}{c}
\min _{\boldsymbol{\alpha} \in R^{(l+u)}, \boldsymbol{\xi}\in R^{l}} \frac{1}{l} \sum_{i=1}^{l} \xi_{i}+\lambda_{1} \boldsymbol{\alpha}^{T} K \boldsymbol{\alpha}+\lambda_{2} \boldsymbol{\alpha}^{T} K \bar{L}^{s *} K \boldsymbol{\alpha} \\
s.t. y_{i}^{s l}\left(\sum_{j=1}^{l+u} \alpha_{j} k\left(x_{i}^{s}, x_{j}^{s}\right)+b\right) \geq 1-\xi_{i}, i=1, \ldots, l \\
\xi_{i} \geq 0, i=1, \ldots, l
\end{array}
\end{equation}
where $b$ is the bias term, $\xi_i$ is the slack variable.

Equation~\ref{19} can be effectively solved using the Lagrange dual~\cite{belkin2006manifold}. Therefore, we can obtain:
\begin{equation}\label{20}
\begin{aligned}
&\boldsymbol{\beta}=\arg \max _{\boldsymbol{\beta} \in R^{l}} \sum_{i=1}^{n} \beta_{i}-\frac{1}{2} \boldsymbol{\beta}^{T} Q \boldsymbol{\beta}\\
&\text {s.t. } \sum_{i=1}^{l} \beta_{i} y_{i}^{s l}=0,0 \leq \beta_{i} \leq \frac{1}{l}, i=1, \ldots, l
\end{aligned}
\end{equation}
where $\boldsymbol{\beta}$ is Lagrange multipliers matrix,  $Q=\hat{Y}\hat{J}K\left(2\lambda_1I+2\lambda_2{\bar{L}}^{s\ast}K\right)^{-1}{\hat{J}}^T\hat{Y}$, $\hat{Y}=diag\left(y_1^{sl},...,y_l^{sl}\right)$, $\hat{J}=\left[I_l,\mathbf{0}\right]\in R^{l\times\left(l+u\right)}$ , and $I_l\in R^{l\times l}$ is the identity matrix.

The adaptive SVM classifier can be easily constructed by using a standard SVM solver with the quadratic form induced by $Q$. Then the obtained $\boldsymbol{\beta}$ is used to solve the parameter $\boldsymbol{\alpha}$ by $\boldsymbol{\alpha}=\left(2 \lambda_{1} I+2 \lambda_{2} \bar{L}^{s *} K\right)^{-1} \tilde{J}^{T} \hat{Y} \boldsymbol{\beta}$.

The complete procedure of DGL with hinge or squared loss is summarized in Algorithm~\ref{alg_1}.
\begin{algorithm}[htb]
  \caption{ Adaptive cross-domain classifier construction}
  \label{alg_1}
  \renewcommand{\algorithmicrequire}{\textbf{Input:}}
  \renewcommand{\algorithmicensure}{\textbf{Output:}}
  \begin{algorithmic}[1]
    \Require
    Source and target data $X_s$, $X_t$; Parameter:$\ \lambda_1$,$\ \lambda_2$, $\xi$; Gaussian kernel function $k$.
 \State Learning a domain-invariant graph:
 \State Compute $L^s$, $L^t$, and $L^{st}$;
 \State Achieve the eigensystem $\left\{\Phi^t,\Lambda^t\right\}$ of $L^t$;
 \State  Estimate the eigensystem ${\bar{\Phi}}^s$ of  ${\bar{L}}^{s*}$ by \ref{7};
 \State Solve optimal problem~\ref{10} for $\Lambda^*$;
 \State Adaptive semi-supervised classifier:
 \State Plug ${\bar{L}}^{s*}={\bar{\Phi}}^s\Lambda^*\left({\bar{\Phi}}^s\right)^T$ into~\ref{15};
 \State Use the kernel function to construct Gram matrix $K$ \State  Compute $\boldsymbol{\alpha}$ for adaptive RLS classifier by Eq.~\ref{18} and for adaptive SVM classifier by Eq.~\ref{20}
    \Ensure
      An adaptive classifier $f$;
  \end{algorithmic}
\end{algorithm}

\subsection{ Complexity Analysis}

The computational complexity of the model includes three parts.

 For build ${\bar{L}}^{s\ast}$, computing graph Laplacian costs $\mathcal{O}\left( m\left( n_s+n_t \right) ^2 \right)$, achieving the eigensystem of $L^t$ costs $\mathcal{O}\left( \left( n_t \right) ^3 \right)$, estimating eigenvectors of ${\bar{L}}^{s\ast}$ costs $\mathcal{O}\left( n_s\left( n_t \right) ^2 \right)$, and solving optimal problem for $\Lambda^\ast$ costs $\mathcal{O}\left( 2\left( n_t \right) ^3 \right)$.

Solving the objective 14 using LU decomposition costs $O\left( \left( n_s \right) ^3 \right)$. Solving the SVM optimization~\ref{20} with the SVM solver~\cite{SVM} costs $\mathcal{O}\left( \left( n_s \right) ^{2.3} \right)$. For constructing the kernel matrix $K$ costs $\mathcal{O}\left( m\left( n_s+n_t \right) ^2 \right)$.

Therefore, in summary, the computational complexity of Algorithm~\ref{alg_1} is $\mathcal{O}\left(m\left(n_{s}+n_{t}\right)^{2}+\left(n_{t}\right)^{3}+n_{s}\left(n_{t}\right)^{2}+\left(n_{s}\right)^{3}\right)$.

\subsection{ Connections to Prior Works}
We roughly explained the differences between the proposed DGL and other existing methods in Section~\ref{section2}. DGL is substantially different from the prior proposed domain adaptation methods, such as \cite{DBLP:conf/cvpr/GongSSG12,long2015domain,wang2018visual}. All these methods rely on large amounts of source label information, which limits their application in real situations. In this subsection, we specifically discuss the differences between DGL and other methods from an insightful perspective.

\begin{enumerate}
  \item Distribution mismatch reduction: The distribution divergence between domains is the fundamental challenge to generalize the source model to the target domain. Therefore, formalizing the distribution mismatch and designing useful strategies to minimize it is the most crucial thing in domain adaptation. Maximum Mean Discrepancy (MMD)\cite{DBLP:conf/nips/GrettonBRSS06}, as a nonparametric measurement, has been widely used in many domain adaptation methods to compare the distribution distance\cite{long2014adaptation,wang2018visual}. However, MMD highly depends on the nonlinear kernel map, and the kernel function may be not optimal for different transfer situations.
  \item Scope of application: Almost all the prior methods are feasible with a precondition, i.e., enough source label information is available. Although we have rich data resources in practical applications, the label information is seriously insufficient. From this perspective, it is difficult to promote the methods that rely on large amounts of labeled data in practical applications.
\end{enumerate}

DGL is completely different from the prior methods.

\begin{enumerate}
  \item DGL flexibly uses the Nystr$\rm\ddot{o}$m approximation error to formalize the distribution mismatch in a convenient way. DGL has the different perspectives to analyze the connection between domains and finally uses the learnt domain-invariant graph to link two domains geometrically.
  \item Different from prior methods, the implementation process of DGL does not need to rely on a large amount of label information of source data, thereby broadening the scope of application of DGL. Even when the label information is sufficient, DGL can still achieve comparable accuracy.
\end{enumerate}

In summary, DGL can simultaneously explore the distribution divergence minimization and the challenge of domain adaption, where only a few labeled source data are available. Moreover, DGL is a general idea in which the standard methods can be easily integrated.

\section{Experiment}\label{section5}
We conduct experiments on different kinds of datasets, including Reuters-21578, Office, and Caltech-256, to show the performance of DGL under different amounts of source domain label information.

\subsection{ Data Description}

Reuters-21578 is a challenging text dataset with many categories. There are three top categories including orgs, people, and place. We directly adopt the preprocessed Reuters-21578 dataset by Gao \emph{et al.}~\cite{gao2008knowledge}, and build three transfer tasks including ``org vs people'', ``people vs place'' and ``org vs place''.

Office and Caltech-256 are are two standard image datasets widely used  to evaluate the domain adaptation methods. Office~\cite{li2014learning,liu2019structure-preserved} comprises of three object domains, Amazon (A), Webcam (W), and DSLR (D). Amazon contains image downloaded from online merchants, Webcam includes low-resolution images, and DSLR consists of high-resolution images. There are totally 4652 images and 31 categories. Caltech-256~\cite{article1} includes 30607 images and 256 categories. We use the public Office+Caltech dataset with SURF features released by Gong \emph{et al.}~\cite{DBLP:conf/cvpr/GongSSG12} in these experiments. Finally, we build twelve transfer tasks, e.g., ``A vs C'', ``A vs D'', ``A vs W''.

\subsection{ Comparison Methods}
To evaluate the performance of DGL, we choose several competitive domain adaptation algorithms to compare with DGL on different transfer tasks.

\begin{itemize}

\item Regularized Least Squares (RLS)
\item Support Vector Machine (SVM)
\item Transfer Component Analysis (TCA) ~\cite{Sinno2011Domain}
\item Geodesic Flow Kernel (GFK) ~\cite{DBLP:conf/cvpr/GongSSG12}
\item Transfer Kernel Learning (TKL) ~\cite{long2015domain}
\item Transfer Joint Matching (TJM) ~\cite{long2014transfer}
\item Joint Geometrical and Statistical Alignment (JGSA) ~\cite{zhang2017joint}
\item Balanced Distribution Adaptation (BDA) ~\cite{wang2017balanced}
\item Guide Subspace Learning (GSL) ~\cite{zhang2019guide}
\end{itemize}

RLS and SVM are two standard learning algorithms, which can be treated as the special cases of DGL\_RLS and DGL\_SVM, i.e., $\lambda_2=0$. TCA, GFK, TKL, TJM, JGSA, BDA, and GSL are competitive domain adaptation frameworks. Specifically, TCA aims to find a shared subspace by minimizing maximum mean discrepancy across domains. TJM aims to construct new data representations by matching the features and reweighting data across domains. JGSA learns two projections that project the domain data into the subspaces where the geometrical and distribution shifts are reduced. BDA aims to build a shared subspace via minimizing the marginal and conditional distribution discrepancies adaptively. GSL proposes to learn an invariant and discriminative domain agnostic subspace where the distribution mismatch can be minimized.
\subsection{ Experiment Setting}
We randomly sample 5\%, 10\%, 25\%, 50\% of source samples as labeled source samples, and the remaining samples are unlabeled. No label information is available in the target domain. For all the methods, we will conduct these methods based on different percents of labeled source samples and test them in the target domain. All experiments are repeated five runs, and we finally take the average of the experimental results.

The parameters of all comparison methods are set as suggested in their original papers. In DGL, there are three main parameters, i.e., $\lambda_1$, $\lambda_2$ and $\xi$. We tune these parameters in the case of 5$\%$ labeled source samples. Specifically, $\lambda_1$ and $\lambda_2$ are tuned in a range $\left\{0.001,0.005,0.01,0.05,0.1,0.5,1,5,10\right\}$. $\xi$ is tuned in a range $\left\{1,\ 1.1,1.2,1.3,1.5,2,2.5,3,5\right\}$. The best results are reported, for image datasets $\lambda_1=10$, $\lambda_2=0.001$, $\xi=1$ and for text datasets $\lambda_1=5$, $\lambda_2=0.001$,$\ \xi=2$.

We utilize classification accuracy on target data as the evaluation metric~\cite{duan2012domain,jiang2019stacked}.
\begin{equation}\label{21}
\begin{split}
Accuracy(\%)=\frac{\left| x_i\in X_t\land f\left( x_i \right) =y\left( x_i \right) \right|}{\left| x_i\in X_t \right|}\times 100
\end{split}
\end{equation}
where $f\left(x_i\right)$ is the prediction label of target sample $x_i$, $y\left(x_i\right)$ is the ground truth label of $x_i$.

\subsection{ Experiment Results and Analysis}
We illustrate the experimental results and provide a brief analysis of them.

The experiment results are shown in Table~\ref{tab:addlabel}-Table~\ref{tab:addlabe4}. Each Table shows the average accuracy of the experiment results in different transfer tasks with different label rates. The best experimental results are shown in bold. Moreover, in order to further observe the effectiveness of DGL. We build another seven transfer tasks where only 5\% source label information can be available. From these results, we can make the following analysis.
\begin{figure}
\begin{center}
\includegraphics[width=3in]{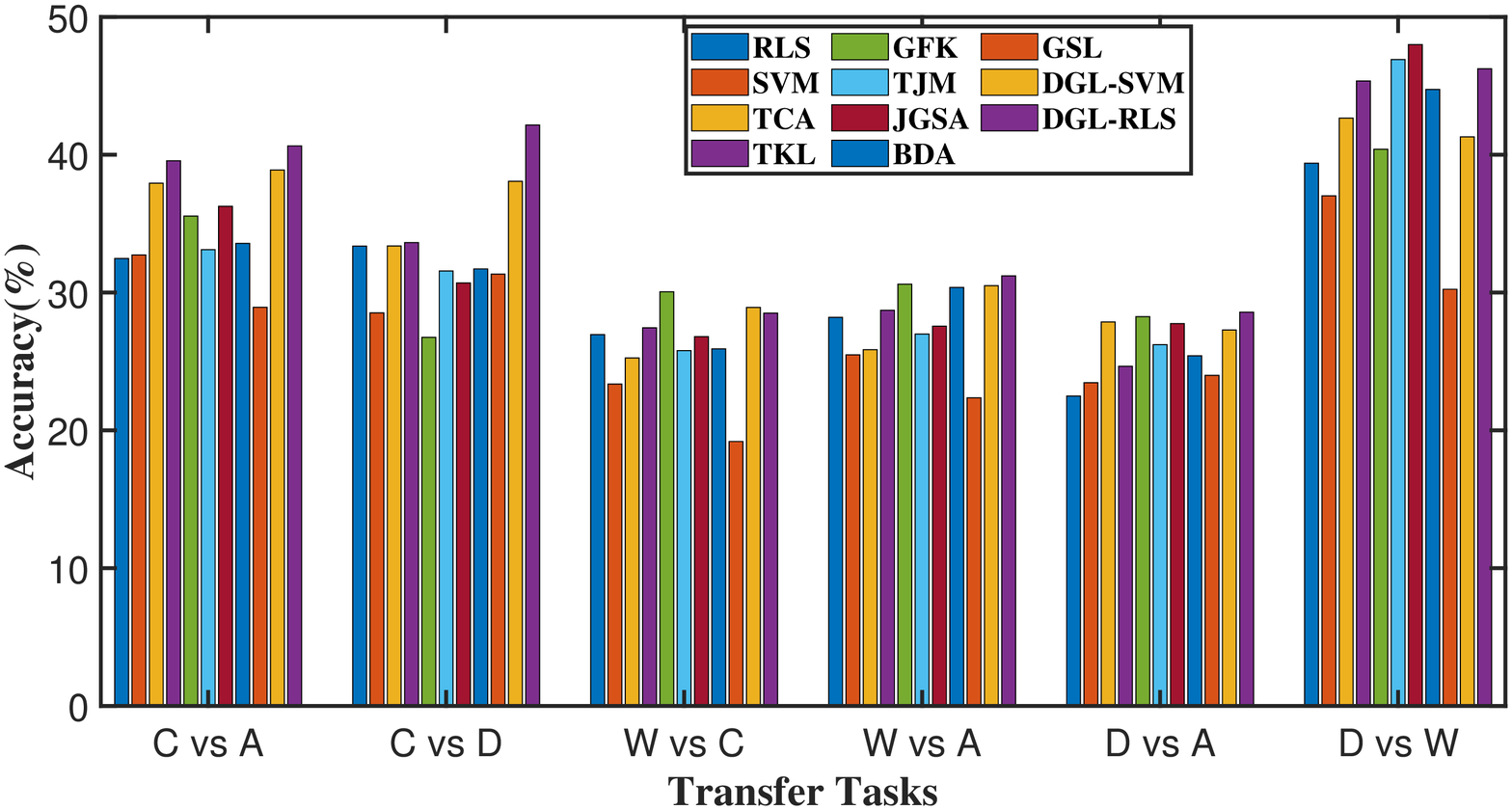}
\end{center}
   \caption{Average classification accuracy (\%) on the seven transfer tasks with 5\% labeled source data}
\label{figure2}
\end{figure}

Firstly, from Table~\ref{tab:addlabel}, we observe that TCA, TKL, GFK, TJM, JGSA, BDA, and GSL these domain methods fail to predict target data in all of the transfer tasks when 5\% labeled source data are available during transfer process. A major limitation of these methods is that they rely on large amounts of source label information. In real applications, it is time-consuming and labor-costing to collect large amounts of labeled source samples and thus results in their poor performance when there is scarce label information existing in the source domain. However, DGL can ensure classification accuracy with a few labeled source samples. That is because the domain-invariant graph enables the classifier constructed on source domain to be adaptive to target domain. At the same time, this invariant graph enables DGL to employ unlabeled source samples to assist knowledge transfer, thereby reducing reliance on source label information.

Secondly, from the Table~\ref{tab:addlabel}, DGL achieves much better  performance than the baseline methods. we can  find that the average classification accuracy of DGL on the different transfer tasks is 46.45\%. The accuracy improves 3.84\% compared to the best baseline method GFK. Similarly, from the Table 2 to Table 3, the average classification accuracy of DGL with different source label rate is 48.75\% and 50.69\% respectively. The accuracy improves 1.24\% and 1.12\% compared to the best baseline method TKL. From the Table 4, with the increase of labeled source data, DGL can still achieve comparable performance than GSL.

Thirdly, the standard RLS and SVM classifiers perform poorly on many transfer tasks. Both methods treat the data from different domains as if they are drawn from the identical distribution. However, this strict assumption is invalid in the real world and therefore results in their poor performance in cross-domain classification problems.

Further analysis of DGL can be observed in Fig.~\ref{figure2}. We build another six transfer tasks where only 5\% source label information can be available to analyze the effectiveness of DGL further. From Fig.~\ref{figure2}, we can find that DGL outperforms the baseline methods in most cases. Moreover, we can observe that in some transfer tasks, the traditional learning method RLS or SVM performs better than some comparison domain adaptation methods. For example, in transfer tasks ¡°C vs D¡±, RLS performs better than TJM, JGSA, BDA, and GSL. We analyze that when there is not enough source label information, implementing knowledge transfer across domains compulsorily will lead to a severe decline in algorithm performance, also known as the negative transfer. However, DGL can achieve stable performance even though there is scarce source label information. Since the analysis is concluded from a large number of experiments, it convincingly demonstrates that DGL can build a robust adaptive classifier for the target domain, especially when there is scarce source label information.

\begin{figure}
\begin{center}
\includegraphics[width=3in]{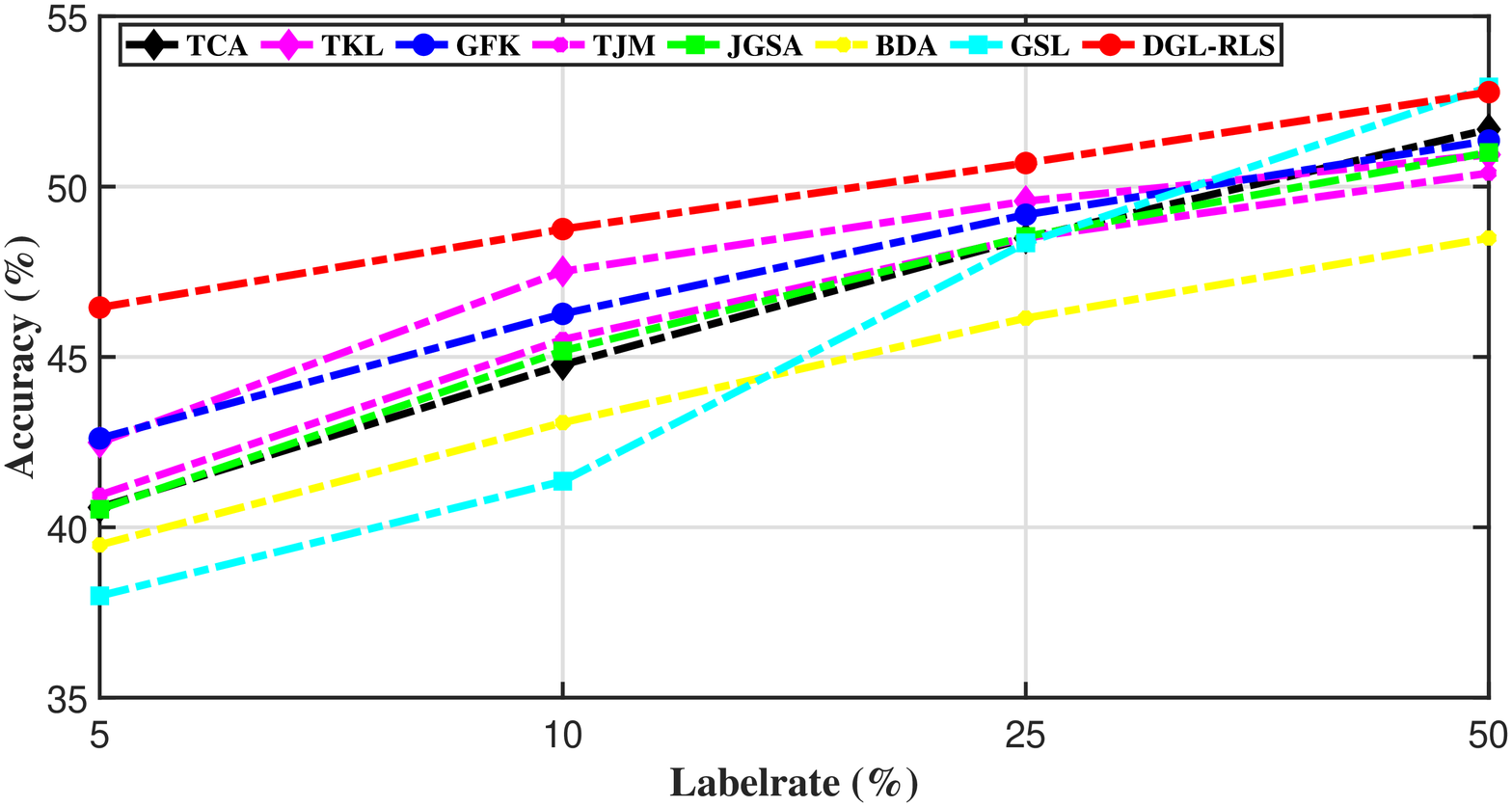}
\end{center}
   \caption{Performance of different methods with different labelrate}
\label{figure3}
\end{figure}

In order to further analyze the influence of source label information in different transfer learning methods, we illustrate the accuracy of different methods with different source label information, referring to Fig.~\ref{figure3}. From Fig.~\ref{figure3}, we find that the performance of all domain adaptation methods is growing better with increasing source label information. However, we observe that the performance ``TKL'' and ``GSL'' increases steeper than the other methods. This shows that these algorithms are more dependent on the source domain label information. We can also find that DGL increases stably with the growing label information, which shows that DGL has strong adaptability and stability. That means DGL can implement knowledge transfer without relying on much source label information and can also achieve comparable performance than the baseline method with increasing label information.

However, from the trend in the Fig.~\ref{figure3}, we  speculate that when the source domain label rate reaches 100\%, the performance of DGL is not necessarily significantly better than other methods. We think that when source domain label information is sufficient, merely considering the geometric connection between domains is not enough.  Maybe some statistical information, e.g., means, the variance should also be adapted between domains. We will analyze it in our future work, because the fundamental point of DGL is to solve the problem of domain adaptation with only scarce source label information. From all experimental results, we can observe that DGL can successfully implement knowledge transfer across domains with scarce source label information.

\begin{table*}[htbp]
  \centering
  \caption{Average classification accuracy (\%) on the different transfer tasks with 5\% labeled source data}
    \begin{tabular}{cccccccccccc}
    \toprule
    5\%   & RLS   & SVM   & TCA   & TKL   & GFK   & TJM   & JGSA  & BDA   & GSL   & DGL\_SVM & DGL\_RLS \\
    \midrule
    A vs C & 26.80 & 29.33 & 24.24 & 30.15 & 29.65 & 26.54 & 30.35 & 25.41 & 23.69 & 30.13 & \textbf{30.69} \\
    W vs D & 32.48 & 34.65 & 35.03 & 45.10 & 41.53 & 42.04 & 44.85 & 44.59 & 45.22 & 48.79 & \textbf{53.87} \\
    A vs D & 27.26 & 28.03 & 28.15 & 28.79 & 29.94 & 28.79 & 27.26 & 27.90 & 30.57 & 31.97 & \textbf{33.12} \\
    A vs W & 28.61 & 32.48 & 23.19 & 29.42 & 30.78 & 27.73 & 26.37 & 26.78 & 25.42 & 31.26 & \textbf{31.73} \\
    C vs W & 25.76 & 26.03 & 23.52 & 27.80 & 26.78 & 27.12 & 27.59 & 25.02 & 25.08 & 29.35 & \textbf{33.97} \\
    D vs C & 19.16 & 19.98 & 23.76 & 23.74 & 22.71 & 21.85 & 21.54 & 21.18 & 18.52 & 21.26 & \textbf{24.82} \\
    \midrule
    org vs people & 73.63 & 74.01 & 76.44 & 74.50 & 74.40 & 74.36 & 68.79 & 70.56 & 59.44 & \textbf{77.68} & 75.39 \\
    people vs place & 57.47 & 58.78 & 61.32 & 56.81 & 58.53 & 58.37 & 56.08 & 54.60 & 57.66 & 62.53 & \textbf{64.15} \\
    org vs place & 63.28 & 64.18 & 69.59 & 66.10 & 69.16 & 61.68 & 61.94 & 59.33 & 56.28 & 70.04 & \textbf{70.31} \\
    \midrule
    Average & 39.38 & 40.83 & 40.58 & 42.49 & 42.61 & 40.94 & 40.53 & 39.48 & 37.99 & 44.78 & \textbf{46.45} \\
    \bottomrule
    \end{tabular}%
  \label{tab:addlabel}%
\end{table*}%

\begin{table*}[htbp]
  \centering
  \caption{Average classification accuracy (\%) on the different transfer tasks with 10\% labeled source data}
    \begin{tabular}{cccccccccccc}
    \toprule
    10\%  & RLS   & SVM   & TCA   & TKL   & GFK   & TJM   & JGSA  & BDA   & GSL   & DGL\_SVM & DGL\_RLS \\
    \midrule
    A vs C & 31.70 & 31.81 & 31.15 & 34.37 & 31.74 & 30.61 & 32.16 & 29.14 & 28.58 & \textbf{35.01} & 34.73 \\
    W vs D & 52.87 & 53.00 & 43.06 & 56.69 & 56.56 & 57.58 & 57.20 & 57.63 & 45.86 & 58.34 & \textbf{59.11} \\
    A vs D & 30.06 & 30.45 & 30.32 & 32.74 & 32.48 & 31.75 & 28.41 & 28.92 & 33.76 & 34.27 & \textbf{35.67} \\
    A vs W & 29.83 & 30.58 & 29.56 & 32.81 & 32.14 & 32.20 & 30.10 & 27.66 & 25.76 & 33.69 & \textbf{35.79} \\
    C vs W & 26.58 & 26.78 & 31.59 & 31.59 & 30.85 & 28.27 & 30.20 & 28.00 & \textbf{34.92} & 32.54 & 34.91 \\
    D vs C & 25.08 & 20.91 & 24.38 & 26.56 & \textbf{27.46} & 25.79 & 24.72 & 23.22 & 23.51 & 24.04 & 24.92 \\
    \midrule
    org vs people & 74.85 & 75.50 & 77.10 & 78.24 & 75.55 & 74.58 & 72.16 & 72.68 & 60.02 & \textbf{78.04} & 77.85 \\
    people vs place & 60.50 & 61.34 & 64.23 & 63.53 & 59.54 & 61.16 & \textbf{64.68} & 58.71 & 60.82 & 64.45 & 64.23 \\
    org vs place & 68.30 & 67.96 & 71.52 & 71.07 & 70.07 & 67.66 & 66.96 & 61.68 & 58.96 & 70.15 & \textbf{71.58} \\
    \midrule
    Average & 44.42 & 44.26 & 44.77 & 47.51 & 46.26 & 45.51 & 45.18 & 43.07 & 41.35 & 47.84 & \textbf{48.75} \\
    \bottomrule
    \end{tabular}%
  \label{tab:addlabe2}%
\end{table*}%

\begin{table*}[htbp]
  \centering
  \caption{Average classification accuracy (\%) on the different transfer tasks with 25\% labeled source data}
    \begin{tabular}{cccccccccccc}
    \toprule
    25\%  & RLS   & SVM   & TCA   & TKL   & GFK   & TJM   & JGSA  & BDA   & GSL   & DGL\_SVM & DGL\_RLS \\
    \midrule
    A vs C & 33.78 & 34.26 & 32.93 & 36.96 & 34.73 & 32.34 & 34.48 & 31.77 & 34.28 & \textbf{38.79} & 38.52 \\
    W vs D & 65.22 & 64.71 & 60.00 & 59.74 & 68.79 & \textbf{71.98} & 71.59 & 68.28 & 63.06 & 63.44 & 64.84 \\
    A vs D & 31.85 & 34.39 & 35.41 & 32.99 & 33.21 & 34.01 & 33.63 & 29.04 & 36.31 & \textbf{38.49} & 38.46 \\
    A vs W & 30.98 & 32.00 & 34.31 & 35.32 & 33.70 & 32.34 & 34.24 & 31.67 & 33.90 & 35.26 & \textbf{36.07} \\
    C vs W & 26.98 & 27.25 & 32.40 & 33.90 & 33.02 & 32.54 & 33.10 & 32.14 & 45.42 & 33.54 & \textbf{36.19} \\
    D vs C & 28.98 & 25.38 & 25.56 & \textbf{30.78} & 30.70 & 29.54 & 27.57 & 27.45 & 27.52 & 27.73 & 26.69 \\
    \midrule
    org vs people & 76.97 & 77.40 & 77.90 & 78.89 & 76.14 & 75.28 & 73.87 & 73.95 & 68.71 & \textbf{78.92} & 78.61 \\
    people vs place & 64.12 & 64.20 & 64.85 & \textbf{65.16} & 61.41 & 59.86 & 59.64 & 58.53 & 63.79 & 64.81 & 64.68 \\
    org vs place & 72.20 & 71.18 & 72.98 & 72.41 & 70.87 & 68.55 & 68.68 & 62.42 & 62.13 & 70.20 & \textbf{72.16} \\
    \midrule
    Average & 47.90 & 47.86 & 48.48 & 49.57 & 49.17 & 48.49 & 48.53 & 46.14 & 48.35 & 50.13 & \textbf{50.69} \\
    \bottomrule
    \end{tabular}%
  \label{tab:addlabe3}%
\end{table*}%

\begin{table*}[htbp]
  \centering
  \caption{Average classification accuracy (\%) on the different transfer tasks with 50\% labeled source data}
    \begin{tabular}{cccccccccccc}
    \toprule
    50\%  & RLS   & SVM   & TCA   & TKL   & GFK   & TJM   & JGSA  & BDA   & GSL   & DGL\_SVM & DGL\_RLS \\
    \midrule
    A vs C & 35.62 & 35.81 & 37.19 & 38.24 & 35.63 & 34.68 & 36.92 & 32.05 & 41.23 & \textbf{40.87} & 40.13 \\
    W vs D & 75.16 & 75.16 & 72.48 & 62.17 & 75.67 & 78.09 & \textbf{82.04} & 76.05 & 65.61 & 76.56 & 71.21 \\
    A vs D & 34.26 & 35.26 & 37.71 & 34.01 & 36.31 & 36.69 & 37.83 & 34.01 & \textbf{44.59} & 39.11 & 41.78 \\
    A vs W & 31.80 & 31.39 & 34.85 & 35.73 & 35.05 & 33.83 & 35.86 & 34.71 & \textbf{41.69} & 36.13 & 36.88 \\
    C vs W & 29.69 & 30.71 & 34.44 & 36.68 & 35.93 & 36.07 & 39.05 & 33.19 & \textbf{55.93} & 31.80 & 38.24 \\
    D vs C & 30.67 & 28.69 & 29.83 & 31.76 & 31.22 & 29.71 & 29.81 & 28.03 & 28.76 & \textbf{30.54} & 30.34 \\
    \midrule
    org vs people & 78.33 & 78.14 & 78.43 & 79.30 & 78.07 & 75.63 & 74.34 & 74.89 & 70.53 & \textbf{79.29} & 78.94 \\
    people vs place & 64.27 & 64.25 & 66.31 & \textbf{66.69} & 62.69 & 59.36 & 54.38 & 58.13 & 65.55 & 62.30 & 64.94 \\
    org vs place & 73.59 & 72.52 & 73.88 & 73.87 & 71.59 & 69.51 & 68.77 & 65.36 & 62.42 & 71.15 & \textbf{72.50} \\
    \midrule
    Average & 50.38 & 50.22 & 51.68 & 50.94 & 51.35 & 50.40 & 51.00 & 48.49 & \textbf{52.92} & 51.97 & 52.77 \\
    \bottomrule
    \end{tabular}%
  \label{tab:addlabe4}%
\end{table*}%

\subsection{ Parameter Sensitivity Analysis}
\begin{figure}[!t]
\centering
\subfloat[Classifier complexity controlling parameter $\lambda_1$]{\includegraphics[width=3in]{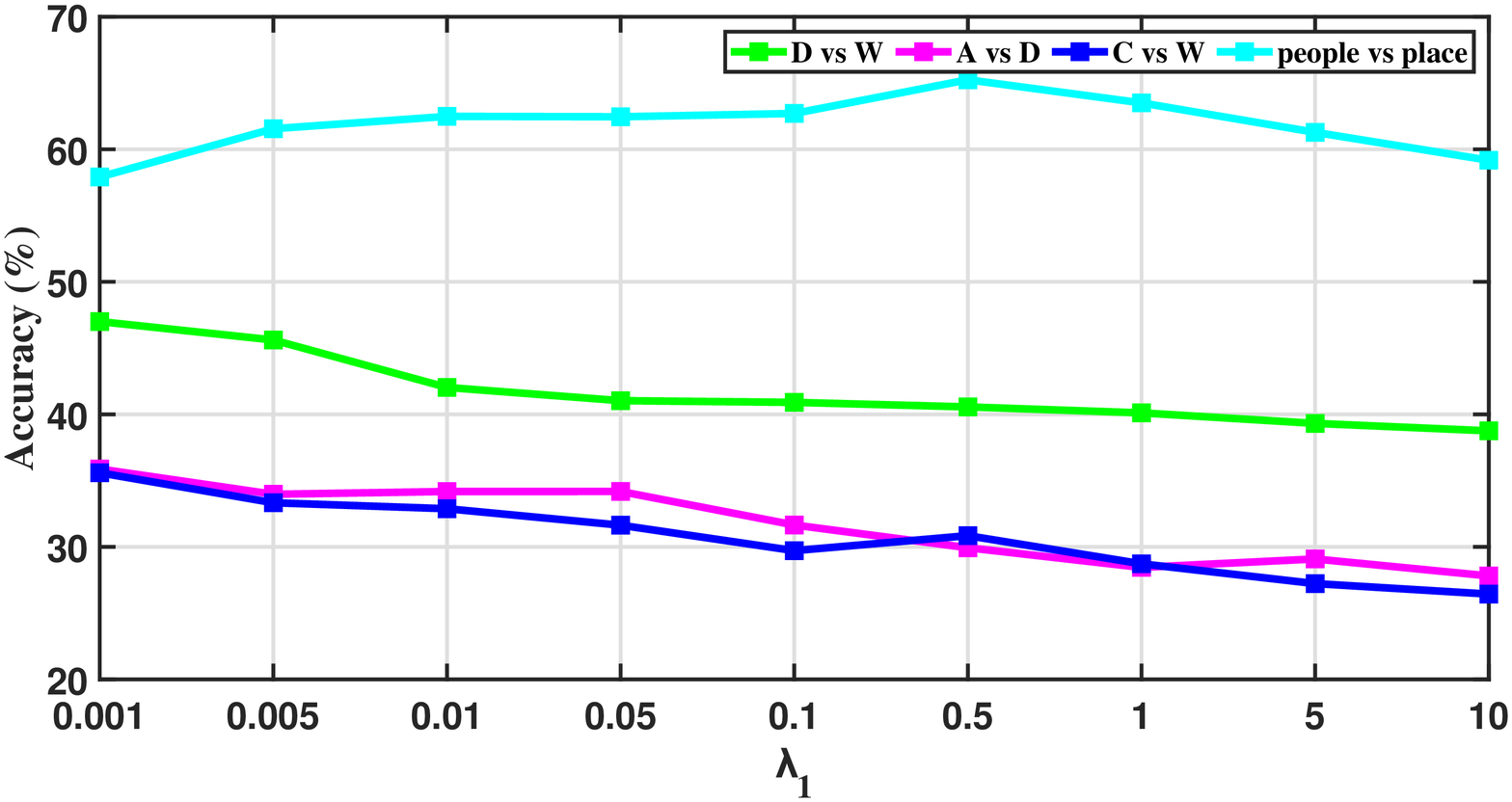}%
\label{fig_fourth_case}}
\hfil
\subfloat[Geometric adaptation controlling parameter $\lambda_2$]{\includegraphics[width=3in]{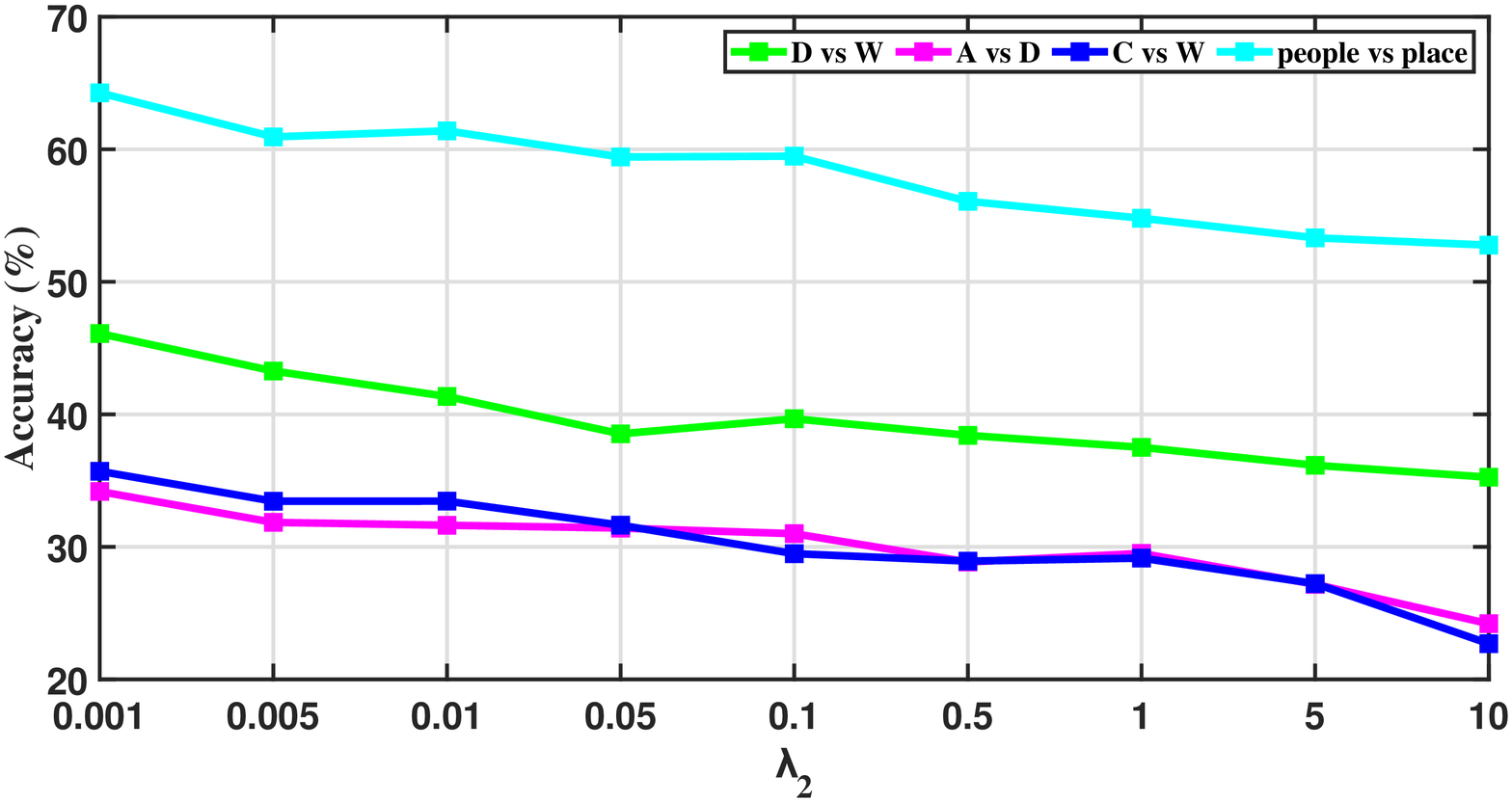}%
\label{fig_fifth_case}}
\hfil
\subfloat[ Damping parameter $\xi$]{\includegraphics[width=3in]{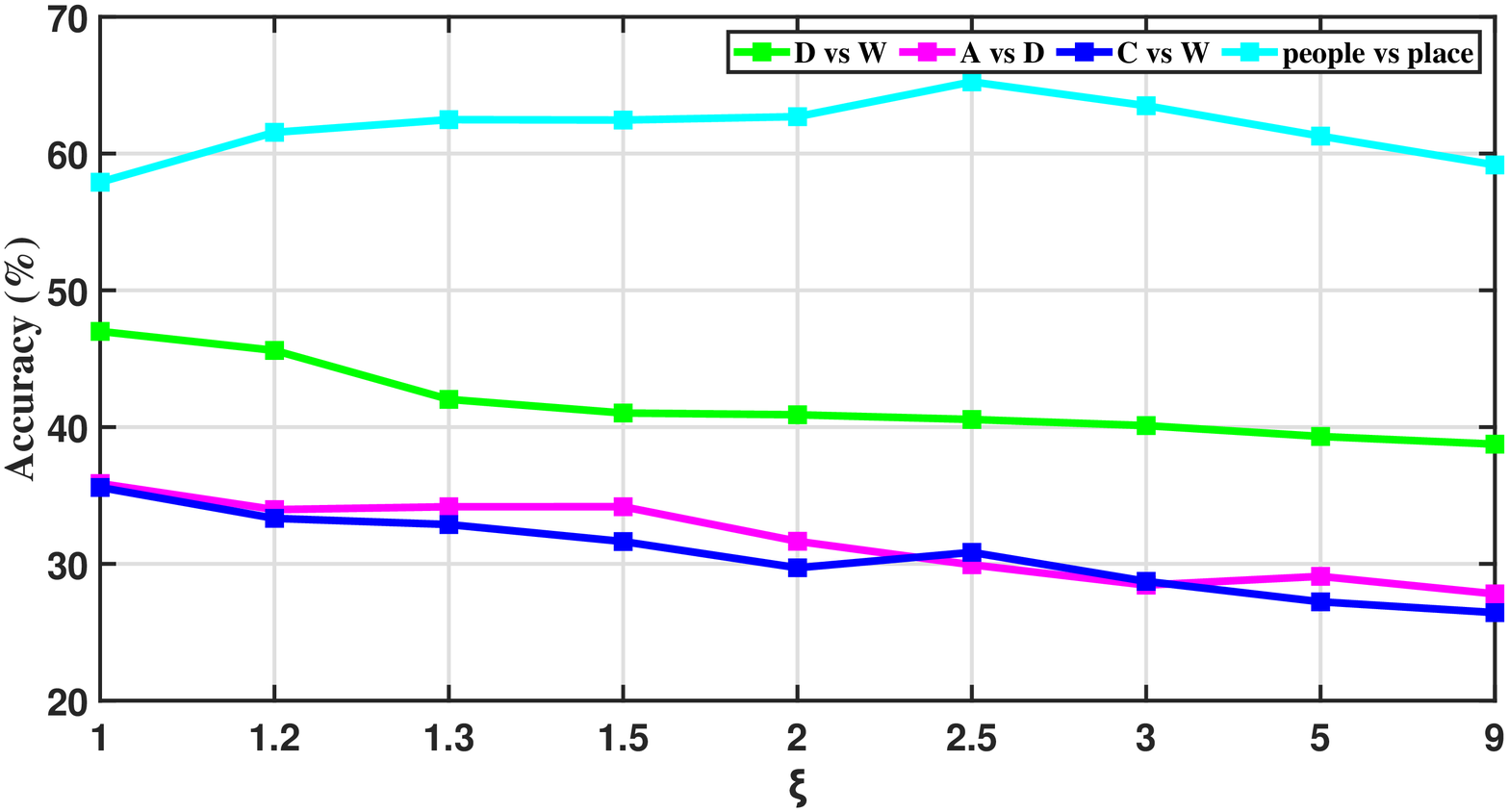}%
\label{fig_twelfth_case}} 
\caption{Parameter sensitivity study for DGL.}
\label{figure4}
\end{figure}

In DGL approach, three parameters should be tuned: classifier complexity controlling parameter $\lambda_1$, geometric adaptation parameter$\lambda_2$ and damping parameter $\xi$. We randomly conduct parameter sensitivity on ``D vs W'', ``A vs D'', ``C vs W'', and ``people vs place'' transfer tasks. 5\% source samples are selected as labeled samples. We implement DGL with the varying values of $\lambda _1\in \left[ 0.001,10 \right]$ by fixing $\lambda_2$ and $\xi$. Similar settings are conducted with $\lambda _2\in \left[ 0.001,10 \right]$  and $\xi\in\left[1,5\right]$. All results are illustrated in Fig.~\ref{figure4} (a)-(c). $\lambda_1$ is used to control the classifier complexity. From the plots, we observe that DGL can perform stably under a wide range $\lambda_1\in\left[1,10\right]$. $\lambda_2$ is used to control the geometric adaptation. We plot classification accuracy with different values of $\lambda_2$. We find that DGL is a little sensitive to the $\lambda_2$, and the best results can be obtained when $\lambda_2=0.001$. $\xi$ can be used to control the decaying trend of the eigen-spectrum. For text dataset, we can choose $\xi\in\left[1.5,2.5\right]$ for stable performance. For image dataset, DGL is a little sensitive to $\xi$, and the best results can be achieved when $\xi=1$.

\section{Conclusion}\label{section6}
In this paper, we proposed a Domain-invariant Graph Learning (DGL) method to cope with a domain adaptation problem where only a few labeled source samples are available. DGL flexibly utilizes the Nystr$\rm\ddot{o}$m approximation error to measure the divergence between the plastic graph ${\bar{L}}^s$ and source graph $L^s$ to formalize the distribution mismatch from the geometric perspective. Through minimizing the approximation error, we construct a domain invariant geometric graph to bridge two domains. Integrating the domain-invariant graph with semi-supervised learning, we further propose an adaptive semi-supervised model, which enables DGL to effectively implement knowledge transfer with scarce source label information. Experimental results on several popular datasets verify that DGL outperforms other comparison methods, especially when there is scarce label information in source domain.

\bibliographystyle{IEEEtran}
\bibliography{bare_jrnl}

\begin{thebibliography}{10}
\providecommand{\url}[1]{#1}
\csname url@samestyle\endcsname
\providecommand{\newblock}{\relax}
\providecommand{\bibinfo}[2]{#2}
\providecommand{\BIBentrySTDinterwordspacing}{\spaceskip=0pt\relax}
\providecommand{\BIBentryALTinterwordstretchfactor}{4}
\providecommand{\BIBentryALTinterwordspacing}{\spaceskip=\fontdimen2\font plus
\BIBentryALTinterwordstretchfactor\fontdimen3\font minus
  \fontdimen4\font\relax}
\providecommand{\BIBforeignlanguage}[2]{{%
\expandafter\ifx\csname l@#1\endcsname\relax
\typeout{** WARNING: IEEEtran.bst: No hyphenation pattern has been}%
\typeout{** loaded for the language `#1'. Using the pattern for}%
\typeout{** the default language instead.}%
\else
\language=\csname l@#1\endcsname
\fi
#2}}
\providecommand{\BIBdecl}{\relax}
\BIBdecl

\bibitem{PanA}
S.~J. Pan and Q.~Yang, ``A survey on transfer learning,'' \emph{IEEE
  Transactions on Knowledge and Data Engineering}, vol.~22, no.~10, pp.
  1345--1359, 2010.

\bibitem{DBLP:journals/corr/abs-1903-04687}
\BIBentryALTinterwordspacing
L.~Zhang, ``Transfer adaptation learning: {A} decade survey,'' \emph{CoRR},
  vol. abs/1903.04687, 2019. [Online]. Available:
  \url{http://arxiv.org/abs/1903.04687}
\BIBentrySTDinterwordspacing

\bibitem{Sinno2011Domain}
S.~J. Pan, I.~W. Tsang, J.~T. Kwok, and Q.~Yang, ``Domain adaptation via
  transfer component analysis,'' \emph{IEEE Transactions on Neural Networks},
  vol.~22, no.~2, pp. 199--210, 2011.

\bibitem{MingshengTransfer}
M.~Long, J.~Wang, G.~Ding, D.~Shen, and Q.~Yang, ``Transfer learning with graph
  co-regularization,'' \emph{IEEE Transactions on Knowledge and Data
  Engineering}, vol.~26, no.~7, pp. 1805--1818, 2014.

\bibitem{YanLearning}
K.~Yan, L.~Kou, , and D.~Zhang, ``Learning domain-invariant subspace using
  domain features and independence maximization,'' \emph{IEEE Transactions on
  Cybernetics}, vol.~48, no.~1, pp. 288--299, 2018.

\bibitem{WeiA}
P.~Wei, Y.~Ke, , and C.~Goh, ``A general domain specific feature transfer
  framework for hybrid domain adaptation,'' \emph{IEEE Transactions on
  Knowledge and Data Engineering}, vol.~31, no.~8, pp. 1440--1451, 2019.

\bibitem{Hua2011Dyadic}
H.~Wang, F.~Nie, H.~Huang, and C.~Ding, ``Dyadic transfer learning for
  cross-domain image classification,'' in \emph{Proceedings of the IEEE
  International Conference on Computer Vision (ICCV)}, 2011, pp. 551--556.

\bibitem{LeiDomain}
L.~Zhang, J.~Yang, and D.~Zhang, ``Domain class consistency based transfer
  learning for image classification across domains,'' \emph{Information
  Sciences}, vol. 418-419, pp. 242--257, 2017.

\bibitem{DingRobust}
Z.~Ding and Y.~Fu, ``Robust transfer metric learning for image
  classification,'' \emph{IEEE Transactions on Image Processing}, vol.~26,
  no.~2, pp. 660--670, 2017.

\bibitem{Zhuang2011Mining}
F.~Zhuang, P.~Luo, Z.~Shen, Q.~He, Y.~Xiong, Z.~Shi, , and H.~Xiong, ``Mining
  distinction and commonality across multiple domains using generative model
  for text classification,'' \emph{IEEE Transactions on Knowledge and Data
  Engineering}, vol.~24, no.~11, pp. 2025--2039, 2012.

\bibitem{article}
R.~Zhao and K.~Mao, ``Supervised adaptive-transfer plsa for cross-domain text
  classification,'' in \emph{Proceedings of the IEEE International Conference
  on Data Mining Workshops (ICDMW)}, 2014, pp. 259--266.

\bibitem{Li2017Knowledge}
Y.~Li, B.~Wei, L.~Yao, H.~Chen, and Z.~Li, ``Knowledge-based document embedding
  for cross-domain text classification,'' in \emph{Proceedings of the
  International Joint Conference on Neural Networks (IJCNN)}, 2017, pp.
  1395--1402.

\bibitem{DBLP:conf/icml/DaiYXY07}
W.~Dai, Q.~Yang, G.~Xue, and Y.~Yu, ``Boosting for transfer learning,'' in
  \emph{Proceedings of the International Conference on Machine learning
  (ICML)}, 2007, pp. 193--200.

\bibitem{DBLP:journals/pami/ChuTC17}
W.~Chu, F.~Torre, and J.~F. Cohn, ``Selective transfer machine for personalized
  facial expression analysis,'' \emph{IEEE Trans. Pattern Anal. Mach. Intell.},
  vol.~39, no.~3, pp. 529--545, 2017.

\bibitem{DBLP:conf/vcip/ChenZL16}
S.~Chen, F.~Zhou, and Q.~Liao, ``Visual domain adaptation using weighted
  subspace alignment,'' in \emph{Proceedings of the Visual Communications and
  Image Processing, (VCIP)}, 2016.

\bibitem{long2015domain}
M.~Long, J.~Wang, J.~Sun, and P.~S. Yu, ``Domain invariant transfer kernel
  learning,'' \emph{IEEE Transactions on Knowledge and Data Engineering},
  vol.~27, no.~6, pp. 1519--1532, 2015.

\bibitem{DBLP:conf/icann/DasL18}
D.~Das and C.~S.~G. Lee, ``Graph matching and pseudo-label guided deep
  unsupervised domain adaptation,'' in \emph{In Proceedings of the
  International Conference on Artificial Neural Networks (ICANN)}, 2018, pp.
  342--352.

\bibitem{DBLP:journals/corr/CourtyFTR15}
N.~Courty, R.~Flamary, D.~Tuia, and A.~Rakotomamonjy, ``Optimal transport for
  domain adaptation,'' \emph{IEEE Transactions on Pattern Analysis and Machine
  Intelligence}, vol.~39, no.~9, 2015.

\bibitem{DBLP:conf/icml/ZhangTK08}
K.~Zhang, I.~W. Tsang, and J.~T. Kwok, ``Improved nystr$\ddot{o}$m low-rank
  approximation and error analysis,'' in \emph{Proceedings of the International
  Conference on Machine learning (ICML)}, 2008, pp. 1232--1239.

\bibitem{belkin2006manifold}
M.Belkin, P.~Niyogi, and V.~Sindhwani, ``Manifold regularization: A geometric
  framework for learning from labeled and unlabeled examples,'' \emph{Journal
  of Machine Learning Research}, vol.~7, pp. 2399--2434, 2006.

\bibitem{DBLP:conf/cvpr/GongSSG12}
B.~Gong, Y.~Shi, F.~Sha, and K.~Grauman, ``Geodesic flow kernel for
  unsupervised domain adaptation,'' in \emph{Proceedings of the IEEE Conference
  on Computer Vision and Pattern Recognition (CVPR)}, 2012, pp. 2066--2073.

\bibitem{DBLP:conf/cvpr/YaoPNLM15}
T.~Yao, Y.~Pan, C.~Ngo, H.~Li, and T.~Mei, ``Semi-supervised domain adaptation
  with subspace learning for visual recognition,'' in \emph{Proceedings of the
  IEEE Conference on Computer Vision and Pattern Recognition (CVPR)}, 2015, pp.
  2142--2150.

\bibitem{wang2018visual}
J.~Wang, W.~Feng, Y.~Chen, H.~Yu, M.~Huang, and P.~S. Yu, ``Visual domain
  adaptation with manifold embedded distribution alignment,'' in \emph{In
  Proceedings of the ACM International Conference on Multimedia (ACMMM)}, 2018.

\bibitem{1262185}
C.~Fowlkes, S.~Belongie, F.~Chung, and J.~Malik, ``Spectral grouping using the
  nystrom method,'' \emph{IEEE Transactions on Pattern Analysis and Machine
  Intelligence}, vol.~26, no.~2, pp. 214--225, 2004.

\bibitem{williams2000using}
C.~Williams and M.~Seeger, ``Using the nystr$\ddot{o}$m method to speed up
  kernel machines,'' in \emph{In Proceedings of Advances in Neural Information
  Processing Systems (NIPS)}, 2000, pp. 682--688.

\bibitem{drineas2005on}
P.~Drineas and M.~Mahoney, ``On the nystr$\ddot{o}$m method for approximating a
  gram matrix for improved kernel-based learning,'' \emph{Journal of Machine
  Learning Research}, vol.~6, pp. 2153--2175, 2005.

\bibitem{li2015large-scale}
M.~Li, W.~Bi, J.~Kwok, and B.~Lu, ``Large-scale nystr$\ddot{o}$m kernel matrix
  approximation using randomized svd,'' \emph{IEEE Transactions on Neural
  Networks and Learning System}, vol.~26, no.~1, pp. 152--164, 2015.

\bibitem{inbook}
M.~Andersen, J.~Dahl, Z.~Liu, and L.~Vandenberghe, \emph{Interior-point methods
  for large-scale cone programming}, 2011.

\bibitem{SVM}
C.~Chih-Chung and L.~Chih-Jen, ``Libsvm: a library for support vector
  machines,'' \emph{ACM Trans. Intell. Syst. Technol., vol. 2(3)}, pp. 1--27,
  01 2011.

\bibitem{DBLP:conf/nips/GrettonBRSS06}
A.~Gretton, K.~M. Borgwardt, M.~J. Rasch, B.~Sch{\"{o}}lkopf, and A.~J. Smola,
  ``A kernel method for the two-sample-problem,'' in \emph{Proceedings of the
  Advances in Neural Information Processing Systems}, 2006, pp. 513--520.

\bibitem{long2014adaptation}
M.~Long, J.~Wang, G.~Ding, S.~J. Pan, and P.~S. Yu, ``Adaptation
  regularization: A general framework for transfer learning,'' \emph{IEEE
  Transactions on Knowledge and Data Engineering}, vol.~26, no.~5, pp.
  1076--1089, 2014.

\bibitem{gao2008knowledge}
J.~Gao, W.~Fan, J.~Jiang, and J.~Han, ``Knowledge transfer via multiple model
  local structure mapping,'' in \emph{The ACM SIGKDD international conference
  on Knowledge discovery and data mining (KDD)}, 2008, pp. 283--291.

\bibitem{li2014learning}
W.Li, L.~Duan, D.~Xu, , and I.~W. Tsang, ``Learning with augmented features for
  supervised and semi-supervised heterogeneous domain adaptation,'' \emph{IEEE
  Transactions on Pattern Analysis and Machine Intelligence}, vol.~36, no.~6,
  pp. 1134--1148, 2014.

\bibitem{liu2019structure-preserved}
H.~Liu, M.~Shao, Z.~Ding, and Y.~Fu, ``Structure-preserved unsupervised domain
  adaptation,'' \emph{IEEE Transactions on Knowledge and Data Engineering},
  vol.~31, no.~4, pp. 799--812, 2019.

\bibitem{article1}
G.~Griffin, A.~Holub, and P.~Perona., ``Caltech-256 object category dataset,''
  \emph{CalTech Report}, 2007.

\bibitem{long2014transfer}
M.~Long, J.~Wang, G.~Ding, J.~Sun, and P.~S. Yu, ``Transfer joint matching for
  unsupervised domain adaptation,'' in \emph{Proceedings of the IEEE Conference
  on Computer Vision and Pattern Recognition (CVPR)}, 2014, pp. 1410--1417.

\bibitem{zhang2017joint}
J.~Zhang, W.~Li, and P.~Ogunbona, ``Joint geometrical and statistical alignment
  for visual domain adaptation,'' in \emph{Proceedings of the IEEE Conference
  on Computer Vision and Pattern Recognition (CVPR)}, 2017, pp. 5150--5158.

\bibitem{wang2017balanced}
J.~Wang, Y.~Chen, S.~Hao, W.~Feng, and Z.~Shen, ``Balanced distribution
  adaptation for transfer learning,'' in \emph{Proceedings of the IEEE
  International Conference on Data Mining (ICDM)}, 2017, pp. 1129--1134.

\bibitem{zhang2019guide}
L.~Zhang, J.~Fu, S.~Wang, D.~Zhang, Z.~Y. Dong, and C.~L.~P. Chen, ``Guide
  subspace learning for unsupervised domain adaptation,'' \emph{IEEE
  Transactions on Neural Networks and Learning Systems}, pp. 1--15, 2019.

\bibitem{duan2012domain}
L.~Duan, I.~Tsang, and D.~Xu, ``Domain transfer multiple kernel learning,''
  \emph{IEEE Transactions on Pattern Analysis and Machine Intelligence},
  vol.~34, no.~3, pp. 465--479, 2012.

\bibitem{jiang2019stacked}
W.~Jiang, H.~Gao, W.~Lu, W.~Liu, F.~Chung, and H.~Huang, ``Stacked robust
  adaptively regularized auto-regressions for domain adaptation,'' \emph{IEEE
  Transactions on Knowledge and Data Engineering}, vol.~31, no.~3, pp.
  561--574, 2019.

\end{thebibliography}
\end{document}